\def\year{2021}\relax
\documentclass[letterpaper]{article} 
\usepackage{aaai21}  
\usepackage{times}  
\usepackage{helvet} 
\usepackage{courier}  
\usepackage[hyphens]{url}  
\usepackage{graphicx} 
\usepackage{natbib}  
\usepackage{caption} 
\usepackage{amsmath,amssymb}
\usepackage{colortbl}
\usepackage[switch]{lineno}

\definecolor{lightGray}{gray}{0.9}
\definecolor{midGray}{gray}{0.7}

\title{Learning Universal Shape Dictionary for Realtime Instance Segmentation}
\author {

        Tutian Tang,
        Wenqiang Xu,
        Ruolin Ye,
        Lixin Yang, 
        Cewu Lu \\
}
\affiliations {
    Shanghai Jiao Tong University \\
    \{tttang, vinjohn, cathyye2000, siriusyang,  lucewu\}@sjtu.edu.cn
}
\begin{document}

\maketitle

\begin{abstract}
We present a novel explicit shape representation for instance segmentation.
  Based on how to model the object shape, current instance segmentation systems can be divided into two categories, implicit and explicit models. The implicit methods, which represent the object mask/contour by intractable network parameters, and produce it through pixel-wise classification, are predominant. However, the explicit methods, which parameterize the shape with simple and explainable models, are less explored. Since the operations to generate the final shape are light-weighted, the explicit methods have a clear speed advantage over implicit methods, which is crucial for real-world applications. The proposed USD-Seg adopts a linear model, sparse coding with dictionary, for object shapes. 
  First, it learns a dictionary from a large collection of shape datasets, making any shape being able to be decomposed into a linear combination through the dictionary.
  Hence the name ``Universal Shape Dictionary''. 
  Then it adds a simple shape vector regression head to ordinary object detector, giving the detector segmentation ability with minimal overhead. 
  For quantitative evaluation, we use both average precision (AP) and the proposed Efficiency of AP (AP$_E$) metric, which intends to also measure the computational consumption of the framework to cater to the requirements of real-world applications. We report experimental results on the challenging COCO dataset, in which our single model on a single Titan Xp GPU achieves 35.8 AP and 27.8 AP$_E$ at 65 fps with YOLOv4 as base detector, 34.1 AP and 28.6 AP$_E$ at 12 fps with FCOS as base detector.
\end{abstract}

\section{Introduction}

Instance segmentation is a fundamental yet challenging task in computer vision. An instance segmentation system should not only detect the location of the desired objects in the images but also produce the object shapes, or delineate the boundary. Mainstream instance segmentation systems adopt either an implicit \cite{maskrcnn,maskScoreRcnn,yolact} or explicit \cite{polarmask,ese} method to model the shapes. \textit{Implicit} methods typically model the shape with intractable parameters in CNNs, and require upsampling modules to produce the mask. Conversely, \textit{explicit} methods model the shape with light-weighted, interpretable representations. Though the implicit methods are predominant in the literature with stronger performance on accuracy generally, the explicit path still draws increasing attention due to the computational efficiency of mask generation. For real-world applications, such as autonomous driving and robot manipulation, accuracy and speed are both important. 

\begin{figure}[t!]
\centering
\includegraphics[width=1\linewidth]{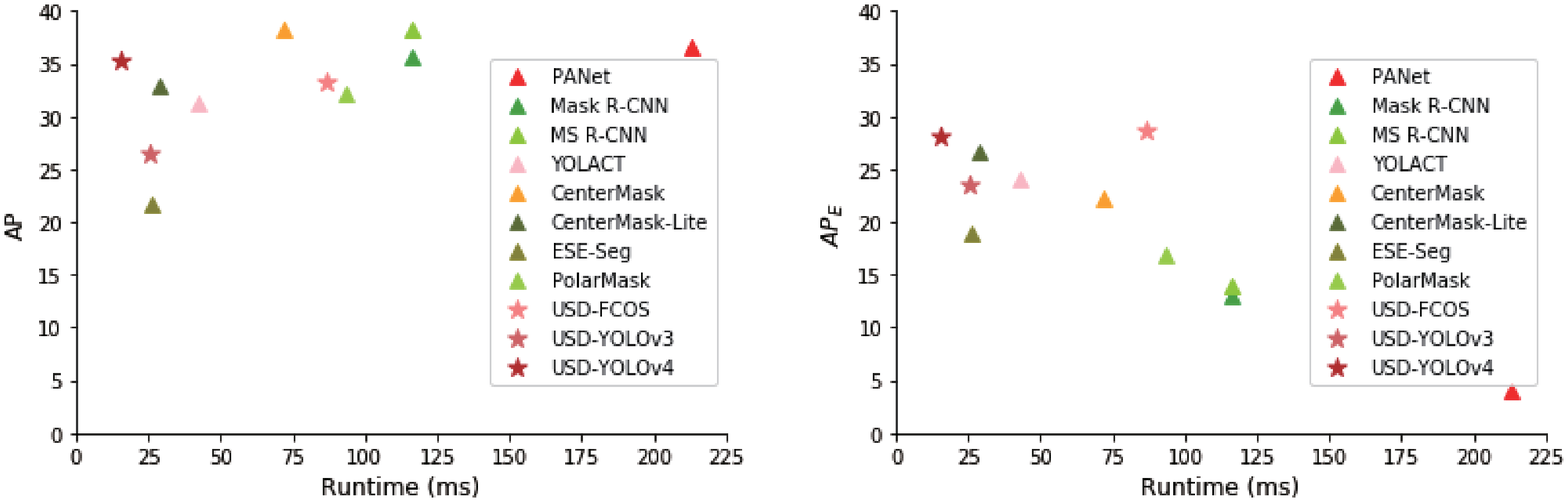}
\caption{The comparison of AP (left) and AP$_E$ (right) on COCO dataset between our method and baselines.}
\label{fig:fig1}
\end{figure}

Inspired by the two sides of requirements, we present a novel instance segmentation method, named \textbf{USD-Seg}. It follows the \textit{explicit} path, representing the shapes by a pre-learned dictionary. Since the dictionary, which is learned from a large collection of shape dataset, e.g. the mask annotation of COCO \cite{coco}, has the generalization ability to unseen objects, it is named ``Universal Shape Dictionary''. We will analyze the representation and generalization ability of the dictionary in \textit{Section Method}. After generating the dictionary, the USD-Seg learns to produce the coefficients to linearly assemble the atoms in the dictionary to obtain the final mask for each object. The whole pipeline adds minimal computational time ($\sim$ 5 ms) upon base object detectors, converts \textit{any} proposal-based detection framework to instance segmentation frameworks effortlessly, and achieves a good balance between accuracy and efficiency.

Furthermore, to quantitatively measure how well the proposed method balances between accuracy and efficiency, we also propose a novel metric, named ``Efficiency Index'' (\textbf{E-Index}), along with the traditional AP. E-Index computes the proportion of the time consumption on the backbone over the whole pipeline. Details of the E-Index will be discussed in \textit{Section Experiment}. Multiplying E-Index with AP forms the novel metric ``Efficiency of AP'' (\textbf{AP$_E$}) for instance segmentation task, which can better reflect both requirements of real-world applications.

We summarize our contributions as follows:
\begin{itemize}
  \item A novel evaluation metric, AP$_E$. It takes a more comprehensive perspective on the instance segmentation task by considering both accuracy and efficiency in a single metric. The E-Index may also be easily applied to other metrics for other tasks like object detection and pose estimation.
  \item A novel instance segmentation approach, USD-Seg. On the challenging COCO dataset \cite{coco}, without any bells and whistles, it achieves state-of-the-art performance on AP metric among all explicit methods and is competitive with implicit methods. When computational efficiency is also taken into consideration, with AP$_E$, the proposed USD-Seg achieves the state-of-the-art among all the methods (See Fig. \ref{fig:fig1}). We hope it can serve as a strong baseline for future works. 
\end{itemize}

\section{Related Works}
\paragraph{\textbf{Instance Segmentation}}Instance segmentation is an important task and many efforts have been put into it. Based on how to produce the mask, researches on instance segmentation can be divided into \textit{implicit} (through neural networks) and \textit{explicit} (through explainable models) methods. 

\textit{Implicit Methods in Instance Segmentation} A majority of the current literature on this field adopts the implicit approach \cite{maskScoreRcnn,maskrcnn,mnc,fcis,spatialEmbed,ssapAffinity,yolact,centermask}, because the complexity of the networks can introduce the redundancy in representation and stabilize the mask generation. 
Some works, like ShapeMask \cite{shapeMask}, obtain a coarse mask with an explicit method in the first stage, and refine it with pixel-wise features in the second stage. Recently, implicit methods like YOLACT \cite{yolact} and CenterMask \cite{centermask} can achieve real-time while maintain good accuracy. 
However, their efficiency comes from the modified feature extraction network, which can be easily exploited by the explicit method as well.

\textit{Explicit Methods in Instance Segmentation} Recently, the explicit approach attracts more and more attention \cite{ese,polarmask}. The simplicity of the mask decoding makes it promising to build up single shot instance segmentation frameworks. ESE-Seg \cite{ese} is currently the fastest instance segmentation framework. PolarMask \cite{polarmask} pushed the accuracy in line with implicit methods. Previous explicit methods generally discuss the polar-coordinated contour representation for object shapes \cite{star_convex,sts,ese,polarmask}. Our proposed approach is the first to explicitly and directly model the mask on pixel level. Obviously, the former representation will fail when representing concave or ring-like shapes (See Fig. \ref{fig:atom_num}).

\paragraph{\textbf{Shape Dictionary Learning}} Researchers have been exploring how to represent the data with some basic elements for many years \cite{onlineDictLearn,dictClass,dictDet13,dictSemseg12,dictSemseg14}. Some general bases are found in the natural images, such as the discrete cosine transform (DCT) \cite{dct}. But, there are also data-driven methods to learn the domain-specific dictionary \cite{onlineDictLearn}. The common way to learn the dictionary is to collect the patches from the natural images and solve an optimization problem that aims to find the dictionary and the coefficients simultaneously. Usually, the generated dictionary has trouble in being generalized to unseen objects. However, as the binary shape mask is relatively simple, the generalization is made possible. Dictionary learning has been used in classification \cite{dictClass}, object detection \cite{dictDet13}, and semantic segmentation \cite{dictSemseg12,dictSemseg14}, but not in instance segmentation.

\begin{figure*}[ht!]
   \begin{center}
      \includegraphics[width=0.9\linewidth]{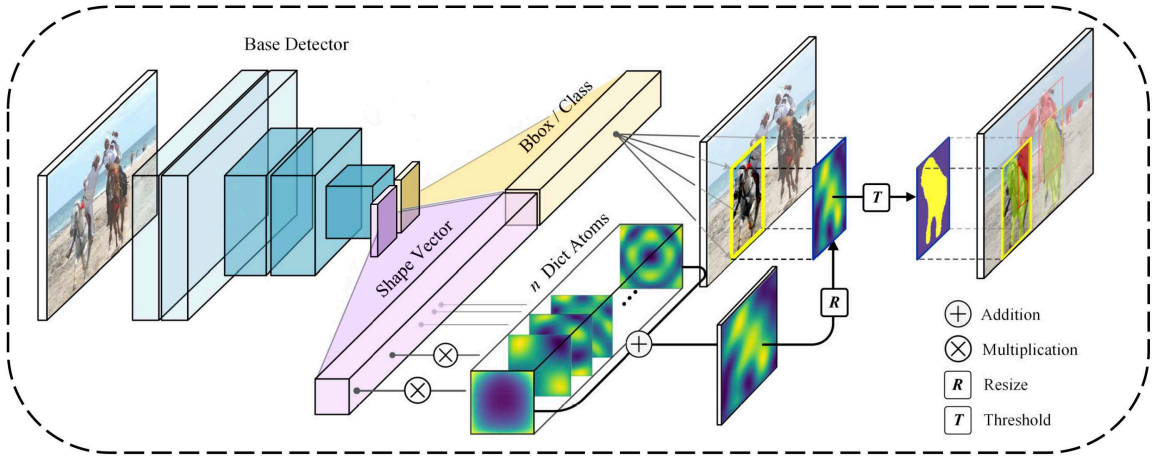}
   \end{center}
      \caption{The overall pipeline of USD-Seg: an RGB image is input to the base detector, and the base detector will regress both detection related information (bounding box and class) and the shape vector. Then the mask will be decoded by simple multiplication between shape vector and dictionary atoms, followed by proper resize and threshold operations. For details of atoms, please refer to Fig. \ref{fig:dict}.}
   \label{fig:framework}
\end{figure*}
\section{Method}
\label{sec:method}

We propose an \textit{explicit}-representation-based instance segmentation framework, which first learns a shape dictionary, and is then integrated into an object detection framework. In fact, very little computation overhead is added, so it can run at almost the same speed as the original detector. The overall pipeline is illustrated in Fig. \ref{fig:framework}.

We will first discuss the generation process and the representation power of the shape dictionary. After that, the shape learning process will be incorporated into the ordinary object detection framework.

\subsection{Generation of Shape Dictionary}
\label{sec:dict}
For each labeled shape in the training dataset, we first resize it to $(W, H)$, and then collect a set of resized object shapes $\mathcal{M}=\{M_{i, c_i}\}_{i=1}^N$, where $N$ is the number of all the shapes, $c_i$ indicates the category of the shape where $c_i\in \{1,\ldots,C\}$, and $C$ is the number of the predefined categories.

The generation of the shape dictionary $D^*$ is to find $n$ atoms $\{d^*_1, \ldots, d^*_n\}$ given $\mathcal{M}$ by optimizing:
\begin{align}\label{equ:dict}
      (U^*, D^*) &= arg\min_{U,D} \frac{1}{2}||M - UD||_2^2 + \lambda ||U||_1 \\ 
      & s.t. \quad ||D_k||^2_2 = 1, \quad \forall 1\leq k\leq n,   
\end{align}
where $M\in \mathbb{R}^{N\times WH}$ is the tensor form of the training shape dataset, $D \in \mathbb{R}^{n\times WH}$ is the dictionary to be optimized, and $U \in \mathbb{R}^{N\times n}$ is the coefficients. $U^*$ and $D^*$ are the learned coefficients and dictionary. $\lambda$ is the balance term to adjust the regularization of $U$. The optimization follows the procedures in \cite{onlineDictLearn}.

After the dictionary is learned, given a shape, unseen or not, its coefficient can be simply obtained by linear regression.
For a single shape, the learned coefficients are denoted $u^*$, which can serve as the ground truth when training if it is from the training set.

\begin{figure}[ht!]
   \begin{center}
      \includegraphics[width=1\linewidth]{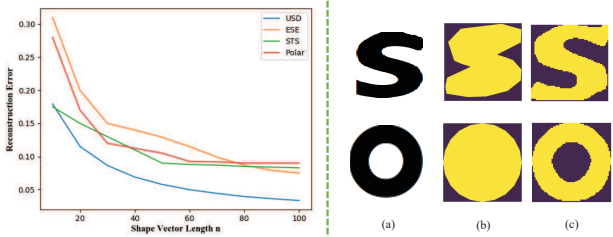}
   \end{center}
      \caption{\textbf{Left:} Reconstruction error with respect to the shape vector length n of Jetley et al. \cite{sts} (STS), ESE-Seg \cite{ese} (ESE), PolarMask \cite{polarmask} (Polar) and ours (USD). \textbf{Right:} Complex shape reconstruction, (a) the original shape, (b) shape reconstructed with polar-coordinated contour-based representation, (c) shape reconstructed with mask-based representation.}
   \label{fig:atom_num}
\end{figure}

\subsection{Representation Power of Shape Dictionary}
\label{sec:reppower}
With a learned dictionary (as visualized in Fig. \ref{fig:framework}), it is necessary to analyze the representation power of shape dictionary before equipping it on the detectors. If not specified, the size for each shape is $(W, H)=(64, 64)$, and the number of atoms in the dictionary $n=32$. The reconstruction error is measured by $1-mIoU$, where $mIOU$ is calculated by averaging the $IoU$ score between reconstructed shapes and ground truth shapes.
\begin{figure}[ht!]
   \begin{center}
      \includegraphics[width=0.95\linewidth]{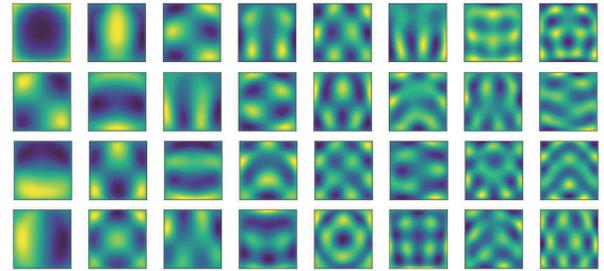}
   \end{center}
      \caption{The atoms of the shape dictionary learned from the COCO train2017 dataset are presented. Yellow stands for high values while blue stands for low values.}
  \label{fig:dict}
\end{figure}

Following previous works \cite{ese}, three aspects of representation power are discussed, namely upper bound of mask accuracy, transferability of the dictionary, and sensitivity to noise. For other aspects such as  consistency in different atom scales, and whether to use class-specific or class-agnostic dictionary, please refer to the supplementary materials.

\textbf{Upper bound of Mask Accuracy}
Assuming the location of object is perfectly accurate with ground truth bounding box, we directly product the off-line estimated $U^*$ and $D^*$ from Eqn. \ref{equ:dict} to obtain the mask $\hat{M}$. Then the resulting $\hat{M}$ should be the upper bound of the predicted mask.
In comparison with former works \cite{sts,polarmask,ese}, the representative power of the USD grows faster and has a lower reconstruction error on the COCO dataset \cite{coco} as the length of the shape vector $n$ grows, as is shown in Fig. \ref{fig:atom_num}. $n$ is the number of the atoms here. It shows our representation performs better on public datasets. To demonstrate the representation power of USD more intuitively, we also illustrate the approximated results for some complex shapes on the right side of Fig. \ref{fig:atom_num} along with the polarized contour-based representation adopted by previous works.

\textbf{Transferability of the Dictionary}
It is also crucial to verify the transferability of the learned dictionary atoms. To be specific, we train them with one dataset and use them to reconstruct object shapes in another dataset. The desirable cross-dataset experimental settings and results are shown in Table \ref{tab:dataset}. The results prove the transferability and the consistent representative power of the learned dictionary.

\begin{figure}[ht!]
   \begin{center}
      \includegraphics[width=0.9\linewidth]{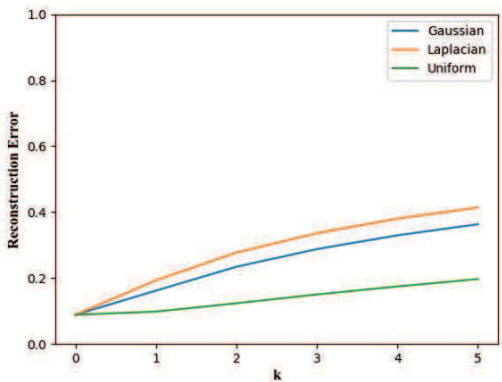}
   \end{center}
      \caption{Sensitivity analysis. Reconstruction error with respect to different degrees of different types of noises.}
   \label{fig:sens}
\end{figure}

\paragraph{\textbf{Sensitivity Analysis of the Dictionary}}
Since the network always produces shape vector with noises, the ability to resist prediction noise should also be examined. Though the noise during training can be modelled by Gaussian distribution or Laplacian distribution according to the loss function, it is intractable to tell the type of noise during testing. Hence, we test sensitivity under some common noise distribution and take the results as reference. The noise distribution experimented are Gaussian ($N(0, 0.05k)$), Laplacian ($La(0,0.05k)$), and uniform ($U(-0.05k, 0.05k)$). The noise is independently added to each coefficient combination, and $k$ is a parameter to adjust the degree of the noise. The reconstruction errors are shown in Fig. \ref{fig:sens}. Overall, this shape vector is robust enough to resist the noise.

\subsection{Object Detection and Shape Learning}
\label{sec:framework}
\textbf{Object Detection Loss}
The object detection loss is the same as the original setting of the base detector, which contains a classification loss $\mathcal{L}_{cls}$ and a bounding box regression loss $\mathcal{L}_{bbox}$, that is:
\begin{equation}\label{equ:det}
   \mathcal{L}_o = \mathcal{L}_{cls}+\mathcal{L}_{bbox}.
\end{equation}

\begin{table}[ht!]
   \centering
   \begin{tabular}{c|c|c|c}
      \hline
      Train & Test &  mIOU & Recon. Err.\\ \hline
      SBD train & SBD val & 0.919 & 0.081\\ 
      COCO train-60 & SBD val & 0.918 & 0.082\\ 
      SBD train & COCO val-60& 0.915 & 0.085\\ 
      COCO train-60 & COCO val-60&  0.917& 0.083\\
      SBD train & COCO val-20 & 0.905 & 0.095\\ 
      COCO train-60 & COCO val-20 &  0.906& 0.095\\ \hline
   \end{tabular}
   \caption{Reconstruction error: training and testing on overlapping and non-overlapping categories. ``COCO train-60'' stands for object shapes selected from COCO 2017 training set with no intersection of the categories to SBD dataset. ``COCO val-20'' stands for the object shapes selected from the validation set of COCO 2017 of the same categories as SBD. The rest can be inferred. In SBD train, 19106 shapes are collected, while in COCO train-60, 367180 shapes are collected.}\label{tab:dataset}
\end{table}

\textbf{Shape Vector Loss}
Due to the sparsity of the dictionary representation, we adopt a cosine similarity based loss rather than $L_{1,2}$ or smooth $L_1$. Given predicted coefficient $\hat{u}_i$ for object $i$, ground truth coefficients $u_i^*$. The cosine similarity loss is given by:

\begin{equation}
   l_i = 1-\frac{\hat{u}_i\cdot u_i^*}{\max\{||\hat{u}_i||_2\cdot||u_i^*||_2, \epsilon\}},
\end{equation}
where $\epsilon=10^{-8}$ is a small value to avoid division by zero.

The overall objective function is
\begin{equation}\label{equ:vec}
  \mathcal{L}_s=\sum_i l_i.
\end{equation}
The quantitative comparison of cosine similarity loss, $L_{1,2}$ loss and smooth $L_1$ loss are in \textit{Section Ablation Study}.


\textbf{Final Objective Function}
Combining Eq. \ref{equ:det} and Eq. \ref{equ:vec}, our final objective function is
\begin{equation}\label{equ:final}
   \mathcal{L} = \lambda_{1}\mathcal{L}_{o}+\lambda_{2}\mathcal{L}_{s},
\end{equation}
where $\lambda_{1}$ and $\lambda_{2}$ are hyper-parameters. 

\textbf{Inference} During inference, the mask will be obtained by $\hat{M} = \hat{u}D^*$. The whole process can be easily implemented in parallel on GPU. Besides, since we did not intentionally control the value range of $U^*$, the predicted $\hat{M}$ may also not be bounded, which makes the binary threshold hard to be determined. To address this numeric issue, we convert all the 0 in the binary shape mask set $\mathcal{M}$ to -1. The whole learning process is based on this converted dataset. Then, the predicted $\hat{M}$ can be easily binarized with threshold $\theta=0$ during inference. For more details, please refer to \textit{Section Ablation Study}.

\section{Experiments}

\subsection{Metric: Efficiency of AP, AP$_E$}
\label{sec:eap}
In this section, we will describe the details of the AP$_E$ metric. Conventionally, FPS and GFLOPs are the two common metrics used to measure the running speed of an instance segmentation system. However, FPS is highly platform-dependent, while GFLOPs may not be proportional to actual running time due to a number of issues such as caching, I/O, hardware optimization etc. \cite{huang2017speed}

A modern instance segmentation framework can be roughly divided into five parts: pre-processing, backbone, neck, head, and post-processing. Such division is concluded from many popular frameworks like mmdetection \cite{mmdetection}, and research literature \cite{maskrcnn,panet,yolov4,huang2017speed}.  
Among them, backbone (e.g, ResNet-50 \cite{resnet}, DarkNet-53 \cite{yolov3}) is the most standard part which is rarely modified by the designers of instance segmentation frameworks, since they are
 originally proposed for classification task. Thus, we find the backbone a good reference to measure the efficiency, and propose the ``Efficiency Index'' (E-Index) based on it.

\textbf{Efficiency Index, E-Index} With fixed system architecture and input size, we denote the time consumption for each procedure as $T_{pre}$, $T_{back}$, $T_{neck}$, $T_{head}$ and $T_{post}$ respectively. Then, the overall time is $T=T_{pre}+T_{back}+T_{neck}+T_{head}+T_{post}$. 
Inspired by Amdahl's law\footnote{https://en.wikipedia.org/wiki/Amdahl's\_law}, parallel portion in the system determines the computational efficiency. As mentioned earlier, the backbone is the most standard part, and is usually implemented in a full parallel manner. 
Thus, a basic version of E-Index is defined as:
\begin{equation}\label{equ:eff}
  \mathcal{E}=\frac{T_{back}}{T}\times 100\%.
\end{equation}

The E-Index implies that methods with higher E-index will enjoy a greater efficiency boost when the backbone is slimmed by techniques like \cite{mobilenet,shufflenet}.
In terms of implementation, we agree that sometimes the border between \emph{head} and \emph{post} may not be very clear, making it hard to measure $T_{head}$ and $T_{post}$ separately. However, it will not become a problem, since we just need to measure the standard $T_{back}$, along with the total $T$, as suggested by Eqn. \ref{equ:eff}.

However, due to historical issues, most classic methods actually report only $T'=T-T_{pre}$ as the inference time. It's not wise to exactly re-implement all the previous works to obtain full $T$. Therefore, to fairly compare with these works, we also define a legacy version of E-Index (denoted as ``$\mathcal{E}_l$''), by substituting $T$ with $T'$ in Eqn. \ref{equ:eff}.

\textbf{AP$_E$} As AP score and E-Index are independent of each other, and both bounded by 0 and 1, we can simply combine these metric by production:
\begin{equation}
  \text{AP}_E = \mathcal{E}\times AP,
\end{equation}
where AP is calculated in the traditional way.

\subsection{Implementation Details}
Since our method does not depend on the base detector, the base detectors for the main experiments are adopted to compare with previous methods. We also adopt YOLOv4 \cite{yolov4} as the base detector to show that our method can be instantly improved when better object detector is available. For these experiments, we keep the same training schedule and hyper-parameters as the corresponding original detectors.

For example, in ablation study, FCOS-ResNet-50-FPN \cite{fcos} is adopted as our base detector and the hyper-parameters stay the same with FCOS and PolarMask \cite{polarmask}. To be specific, we train the network for 90K iterations with stochastic gradient descent (SGD), which can be finished within one day with 4 Titan Xp GPUs. The initial learning rate is 0.01 and mini-batch size is 16. The learning rate is reduced by a factor of 10 at iteration 60K and 80K, respectively. Weight decay and momentum are set as 0.0001 and 0.9, respectively. We initialize our backbone networks with the weights pre-trained on ImageNet \cite{imagenet}. The input images are resized to have their shorter side being 800 and their longer side less or equal to 1333. Multi-scale training is \textbf{not} used if not specified. The AP is reported on COCO \textit{val2017}.


\subsection{Ablation Study}
\label{sec:ablative}

\textbf{Learning Dictionary During v.s. Before Training}
Since the objective function of dictionary learning (Eqn. \ref{equ:dict}) is differentiable, it can be learned either during or before training. 

If the dictionary is learned during training, the network learns the bases and the vector spontaneously guided by the optimization function (Eqn. \ref{equ:dict}), and then the predicted mask is generated by dot production of the bases and coefficients. We apply direct mask loss (binary cross-entropy) to the predicted and ground truth masks.
However, we find that the network is hard to converge when learning dictionary during training, which leads to a 3.2 mAP decrease. The reason behind may be that CNNs behave differently from OMP (Orthogonal Marching Persuit) \cite{omp}, and the noise originated from prediction is not stable. 

Thus, we decide to train the dictionary by OMP with $\lambda=0.1$ before training the network, and then conduct all the following experiments with the fixed dictionary. In this way, the network only learns and predicts the coefficients. Our implementation relies on scikit-learn \cite{sklearn}. For details, please refer to the supplementary material.

\textbf{Number of Atoms} We have discussed the influence of the number of atoms in the off-line dictionary representation (\textit{Section Method}), but whether it can be transited to CNN is non-trivial. Thus, we conduct the on-line CNN learning experiments with a set of atom numbers (8, 16, 24, 32, 48, 64). The results are shown in Table \ref{tab:atom_num}. Interestingly, we find that the AP converges fast when the number of atoms grows. It is probably due to the sparse nature of the dictionary learning, where most information is encoded by the first few coefficients, and the latter coefficients tend to be suppressed by the regression noise. We select 32 as the default number of atoms for main experiments.
\begin{table}[ht!]
\centering
   \begin{tabular}{c|p{1cm}p{1cm}p{1cm}}
      \hline
      \# Atoms & AP &  AP$_{50}$ & AP$_{75}$ \\ \hline
      8  & 25.3 & 48.4 & 24.9 \\ 
      16 & 27.5 & 48.8 & 27.7 \\ 
      24 & 28.6 & 49.2 & 28.4 \\ 
      32 & 30.1 & 51.7 & 30.3 \\ 
      48 & 30.2 & 51.6 & 30.3 \\ 
      64 & 29.9 & 50.4 & 29.7 \\ \hline
   \end{tabular}
   \caption{The influence of atom number towards instance segmentation performance.}\label{tab:atom_num}
\end{table}
\begin{table}[ht!]
  \centering
   \begin{tabular}{c|ccc}
      \hline
      category & DCT  & Shape Priors & USD \\ \hline
      person & 0.132& 0.107 & 0.093\\ 
      bicycle & 0.170& 0.145& 0.116 \\
      car &0.097 & 0.068& 0.056\\ 
      motorcycle & 0.153&0.128&0.108\\ 
      airplane &0.211&0.184& 0.156\\
      all &0.131&0.103&0.085 \\ \hline
   \end{tabular}
   \caption{Reconstruction error on COCO, DCT vs. Shape Priors vs. USD.}\label{tab:prior}
\end{table}

\begin{table*}[ht!]
\centering
   \begin{tabular}{ccc|p{1cm}p{1cm}p{1cm}|p{1cm}p{1cm}p{1cm}}
      \hline
      cosine & $L_2$ & Smooth $L_1$ & AP  & AP$_{50}$ & AP$_{75}$ & AP$^{bbox}$  & AP$^{bbox}_{50}$ & AP$^{bbox}_{75}$ \\ \hline
      \checkmark& & &30.1 & 51.7 & 30.3 & 36.9 & 57.1 & 39.0 \\
      &\checkmark &  &25.4 & 45.1& 24.9& 31.4& 51.3 & 33.5\\ 
      & & \checkmark & 26.8 & 46.2 & 25.6 & 31.7 & 51.8 & 34.8\\\hline
   \end{tabular}
   \caption{Different loss function shape vector regression optimization.}\label{tab:loss}
\end{table*}

\begin{table*}[ht!]
\centering
   \begin{tabular}{c|cp{0.6cm}|ccp{0.5cm}p{0.5cm}p{0.4cm}p{0.4cm}p{0.4cm}}
      \hline
      Method & Backbone & Neck & Input & T&$\mathcal{E}_l$&AP$_E$ & AP & AP$_{50}$ & AP$_{75}$ \\ \hline
      \textit{implicit methods}\\
      PA-Net \cite{panet} & R-50& PAN & (800,1333)& 212.8 &0.11& 4.1 & 36.6 & 58.0 & 39.3 \\ 
      Mask R-CNN \cite{maskrcnn}& R-101 & FPN  & (800,1333)& 116.3 &0.36& 13.0& 35.7 & 58.0 & 37.8\\ 
      MS R-CNN \cite{maskScoreRcnn} & R-101 & FPN & (800,1333)& 116.3 &0.36&13.9& 38.3 & 58.8 & 41.5 \\ 
      YOLACT \cite{yolact}& R-101 & FPN$^*$ & 550&42.7&0.77&24.0 &31.2 & 50.6 & 32.8  \\
      CenterMask \cite{centermask} & R-101 & FPN & (800,)& 72&0.58& 22.3& 38.3& - & - \\
      CenterMask-Lite \cite{centermask} & R-50 & FPN & (600,)& 29 & 0.81 & 26.6 & 32.9 & - & - \\
      CenterMask2-Lite \cite{centermask} & V-39 & FPN & (600,)& 28 & 0.82 & 30.1 & 36.7 & - & - \\ 
       \hline
      \textit{explicit methods}\\\rowcolor{lightGray}
      ESE-YOLOv3-20$^\dagger$ \cite{ese} & D-53 & FPN$^*$ & 416&26.0 & 0.87 & 18.9 & 21.6 &48.7 & 22.4 \\\rowcolor{lightGray}
      USD-YOLOv3-20$^\dagger$ & D-53 & FPN$^*$ & 416& \textbf{25.5} & \textbf{0.89} & \textbf{23.6} & \textbf{26.5} & \textbf{49.2} & \textbf{27.0} \\
      PolarMask-36 \cite{polarmask}& R-101 & FPN$^*$ &(800,1333)& 93.7 &0.52&16.8 &32.1 & 53.7&33.1  \\
      PolarMask-36 \cite{polarmask}& RX-101 & FPN$^*$ &(800,1333)& 93.7 & 0.52&17.2 &32.9 & 55.4&33.8\\
      PolarMask-36 \cite{polarmask}& RX-101-DCN & FPN$^*$ &(800,1333)& 140.2 & 0.64 & 23.2 &36.2 & 59.4&37.7  \\
      USD-FCOS-36 & R-101 & FPN$^*$ &(800,1333)&\textbf{86.6}& \textbf{0.84} & 28.0 & 33.2 & 56.3 & 34.2 \\
      USD-FCOS-36 & RX-101 & FPN$^*$ &(800,1333)& \textbf{86.6} & \textbf{0.84} & \textbf{28.6}& 34.1 & 57.6 & 35.8 \\
      USD-FCOS-36 & RX-101-DCN & FPN$^*$ &(800,1333)& 128.3 & 0.66 &24.4 & \textbf{36.9} &  \textbf{59.6} & \textbf{38.2}\\\rowcolor{lightGray}
      USD-YOLOv4 & CSPD-53 & PAN & 416& \textbf{15.4} & \textbf{0.77} & \textbf{27.8} & 35.8 & 54.9 & 36.3  \\\rowcolor{lightGray}
      USD-YOLOv4 & CSPD-53& PAN & 512& 18.1 & 0.72 & 26.7 & 37.1 & 57.1 & 37.6 \\ \rowcolor{lightGray}
      USD-YOLOv4 & CSPD-53& PAN & 608& 23.2 & 0.69 & 25.8 & \textbf{37.4} & \textbf{58.8} & \textbf{38.8} \\\hline
   \end{tabular}
   \caption{\textbf{Method:} We report the performance on COCO test-dev dataset. $\dagger$ means results from COCO val2017. \textbf{Backbone:} R-50 means ResNet-50 \cite{resnet}. D means DarkNet \cite{yolov3}, RX is ResNeXt \cite{resnext}, DCN is deformable convolution \cite{deformabconv}, V is VoVNet from \cite{centermask} and CSPD is CSPDarknet from \cite{yolov4}; \textbf{Neck:} FPN means the original structure \cite{fpn}, and FPN$^*$ means modified FPN structure. We do not differentiate different modifications here. PAN is from \cite{panet}; \textbf{Input:} (800,1333) means shorter side is at least 800, and longer side is at most 1333. (800,) means shorter side is 800, and the longer side is not limited. 550 means 550 $\times$ 550. The rest can be inferred. The unit for T is ms, and all Ts are measured on Titan Xp for fair comparison.
   }\label{tab:comp_coco}
 \end{table*}

\textbf{Bottleneck of the Method}
Since the philosophy of our method is to decouple instance segmentation to object detection and mask representation, we analyze the bottleneck by substituting these parts with ground truth respectively. 
Based on the experimental setting of \#Atoms=32 in Table \ref{tab:atom_num}, a perfect mask representation will lead to an 8.1 AP gain, a perfect classification will lead to a 25.3 AP gain, and a perfect localization will lead to a 22.6 AP gain. Thus, the accuracy of the whole system strongly relies on the detector. A more advanced detector can significantly improve the segmentation performance, which is confirmed by Table \ref{tab:comp_coco}.

\textbf{Different Dictionary Learning Techniques}
There are other ways to construct the shape dictionary, such as the DCT bases and the shape priors mentioned in ShapeMask \cite{shapeMask} which generate the shape priors by PCA projection and K-Means clustering. For these methods, we calculate the coefficients with OMP \cite{omp} (same as shape dictionary). We list 5 categories from COCO in Table \ref{tab:prior}, and the rest can be found in the supplementary materials. 
Since the number of DCT bases can only be square numbers, here we use 49 as atom number for all these three methods. Namely, the length of shape vector is 49.
Our method outperforms other two consistently over different categories.

\textbf{Cosine Similarity Loss v.s. $L_{1,2}$, Smooth $L_1$ Loss}
$L_{1,2}$, and smooth $L_1$ loss are known to be more sensitive to the precise value of the vector, while cosine similarity only cares about the direction of the vector in the vector space. Besides, the cosine similarity ranges from -1 $\sim$ 1, so the loss ranges from 0 $\sim$ 2, which is fairly balanced with object detection loss (typically $\sim$ 0.8 as it converges). Therefore, the choice of $\lambda_1$ and $\lambda_2$ from Equ. \ref{equ:final} are not sensitive. By default, we set $\lambda_1=\lambda_2=1$ when using cosine similarity loss, and $\lambda_1=1, \lambda_2=0.5$ for Smooth $L_1$ loss. Please refer to supplementary materials for the sensitivity of this hyper-parameter choice.

Quantitatively, the cosine similarity loss is significantly better than $L_p$ based loss functions, as shown in Table \ref{tab:loss}. Another merit of cosine similarity loss is that it has little impact on the bounding box accuracy. The vanilla object detector (i.e., FCOS \cite{fcos}) has AP$^{bbox}$ of 36.7.

\begin{figure*}[ht!]
   \begin{center}
      \includegraphics[width=1\linewidth]{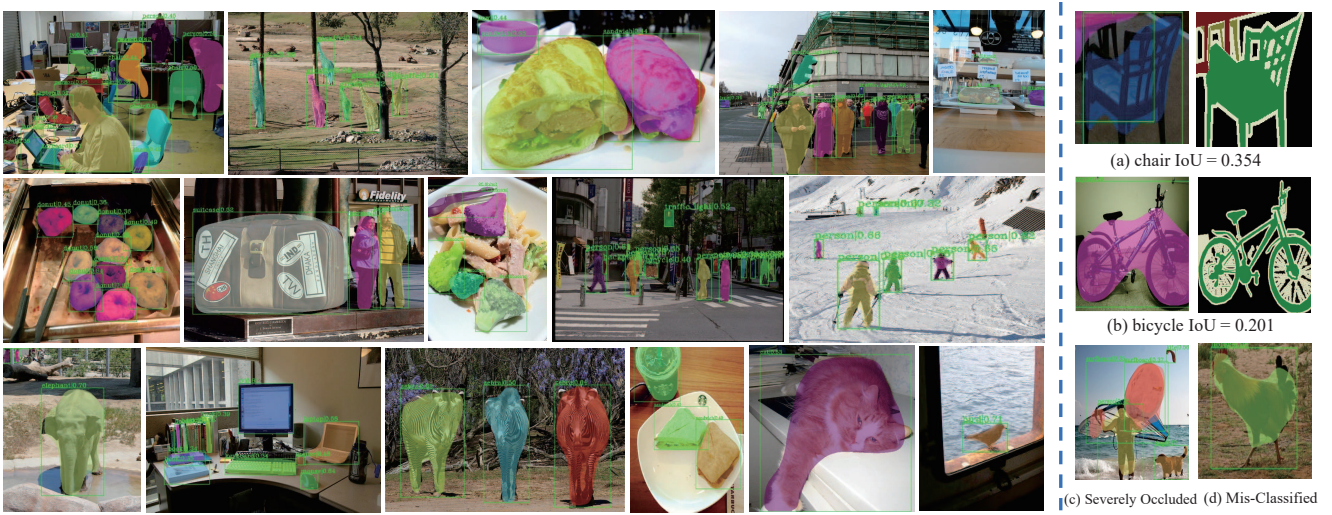}
   \end{center}
      \caption{Qualitative results (left) and typical failures (right).}
   \label{fig:qua_fail}
\end{figure*}
\textbf{Binarized Threshold During Inference} The common practice is to find the minimum ($V_{min}$) and maximum value ($V_{max}$) in a predicted mask, and binarize it according to ($(V_{min}+V_{max})/2$). However, since the value of the predicted masks is unbounded, the common practice will lead to numerical instability. Our experiment verifies it. When we convert the mask from 0/1 binary mask to -1/1 mask, 0 becomes a natural threshold and it gains 0.3 AP stably.

\subsection{Comparison with Other Methods}
\label{sec:comp}
We report the results on COCO \cite{coco} with both standard metrics and our proposed AP$_E$ metric. Results on Pascal 2012 SBD \cite{sbd} are in the supplementary files, which can further confirm the conclusions we draw from the results on COCO.

As shown in Table \ref{tab:comp_coco}, there are three groups reported in the explicit method track. For each group, both the base detector and basic experimental settings are the same as the compared baseline. For example, to compare with ESE-Seg \cite{ese}, model ``ESE-YOLOv3-20'' means ESE-Seg built with Chebyshev polynomial, YOLOv3 as base detector, 20 as the length of shape vector, and $416\times 416$ as input size. Our ``USD-YOLOv3-20'' model adopts the exact same setting except for the shape representation method, namely, the same input size and shape vector length. And since ESE-Seg reports only results from COCO val2017, we also report corresponding results. Similarly, ``USD-FCOS-36'' stands for exact the same setting of PolarMask \cite{polarmask} except for the USD part. It is clear that our method outperforms the previous explicit methods in both speed and accuracy.

Besides, we also compare our method with implicit methods including classic and strong baselines \cite{panet,maskrcnn,maskScoreRcnn} and recent proposed real-time approaches \cite{yolact,centermask}. To note, the ``CenterMask2-Lite'' is not reported from the original paper, but an implementation upgrade after publication. When equipped with YOLOv4 as the base detector, our method outperforms all the real-time methods on AP and all the baselines on AP$_E$. 

When calculating the $\mathcal{E}_l$, we time the process between the processed images sent to the backbone and the predicted masks constructed in the memory. FPN is often modified in some systems \cite{yolact,panet,polarmask}, so we regard it as neck instead of backbone. The naming is in accordance with mmdetection library \cite{mmdetection}.

\subsection{Qualitative Results}
\label{sec:qua}
The qualitative results and typical failure cases are displayed in Fig. \ref{fig:qua_fail}.

\paragraph{\textbf{Failure cases}} We summarize the failure of the shape regression into three different reasons. 1). Detection missing or mis-classification. The quantitative influence was discussed in \textit{Section Ablation Study}. 2). Inconsistent annotation quality. Evaluating on high-quality annotations requires the recovery of the fine details. However, the shape masks annotated by human in the public datasets have inconsistent labeling quality. It makes data-driven method, like dictionary learning, hard to extract the precise high frequency details. 3). Severe occlusion. When there is much overlapping between objects, it becomes hard to be localized and represented by USD.

\section{Conclusion}
In this paper, we propose a novel \textit{explicit}-representation-based instance segmentation framework, which achieves comparable AP performance to implicit pixel-wise classification-based instance segmentation methods, while being significantly faster. The proposed shape dictionary can mitigate two core problems in explicit segmentation, representation and optimization. Besides, the proposed novel metric AP$_E$ can measure the balance between accuracy and efficiency, which may bring a new perspective to instance segmentation task.

\bibliography{usd}

\end{document}